\documentclass[conference]{IEEEtran}
\IEEEoverridecommandlockouts
\usepackage{amsmath,amssymb,amsfonts}
\usepackage{algorithmic}
\usepackage{graphicx}
\usepackage{textcomp}
\usepackage{xcolor}
\usepackage{multirow}
\usepackage{floatrow}
\usepackage[capitalise]{cleveref}
\usepackage{subcaption}
\usepackage[style=ieee, maxnames=6, minnames=1, maxbibnames=6, backend=biber]{biblatex}
\begin{document}

\title{Enhancing Egocentric Object Detection in Static Environments using Graph-based Spatial Anomaly Detection and Correction\\
\thanks{Thanks to Singapore Maritime Institute (SMI)}
}

\author{\IEEEauthorblockN{Anonymous Submission}}
\author{\IEEEauthorblockN{Vishakha Lall}
\IEEEauthorblockA{\textit{Centre of Excellence in Maritime Safety} \\
\textit{Singapore Polytechnic}\\
Singapore \\
vishakha\_lall@sp.edu.sg}
\and
\IEEEauthorblockN{Yisi Liu}
\IEEEauthorblockA{\textit{Centre of Excellence in Maritime Safety} \\
\textit{Singapore Polytechnic}\\
Singapore \\
liu\_yisi@sp.edu.sg}
}

\maketitle

\begin{abstract}
In many real-world applications involving static environments, the spatial layout of objects remains consistent across instances. However, state-of-the-art object detection models often fail to leverage this spatial prior, resulting in inconsistent predictions, missed detections, or misclassifications, particularly in cluttered or occluded scenes. In this work, we propose a graph-based post-processing pipeline that explicitly models the spatial relationships between objects to correct detection anomalies in egocentric frames. Using a graph neural network (GNN) trained on manually annotated data, our model identifies invalid object class labels and predicts corrected class labels based on their neighbourhood context. We evaluate our approach both as a standalone anomaly detection and correction framework and as a post-processing module for standard object detectors such as YOLOv7 and RT-DETR. Experiments demonstrate that incorporating this spatial reasoning significantly improves detection performance, with mAP@50 gains of up to 4\%. This method highlights the potential of leveraging the environment's spatial structure to improve reliability in object detection systems.
\end{abstract}


\section{Introduction}
High-stakes, high-complexity domains such as maritime navigation demand rigorous validation of situational awareness and correct use of critical systems. As maritime training moves toward immersive, high-fidelity simulators, there is a growing need for automated systems that can analyse how users interact with onboard instruments and detect deviations from standard operational behaviour. However, egocentric video captured in these environments presents challenges for conventional computer vision pipelines due to occlusions, varying perspectives, and cluttered layouts.

Egocentric vision has emerged as a rich source of contextual and behavioural cues in fields ranging from robotics to human-computer interaction. Prior work has used wearable cameras to model attention, track tasks, or understand indoor navigation \cite{ego4d, egovideo}. However, few studies address static but semantically dense environments, such as ship bridges, where spatial understanding is crucial and traditional appearance-based models may fail due to perspective distortion or partial views.

This challenge, and the opportunity it presents, is not unique to the maritime domain. In any static environment, object detection performance can be significantly enhanced by incorporating spatial relationships among scene elements, allowing models to reason beyond isolated detections. Graphs provide a powerful abstraction for modelling relationships between objects in visual scenes. Scene graphs have been widely used in image captioning \cite{graph_image_captioning}, visual question answering \cite{graph_question_answering}, and robotics \cite{graph_navigation}. In structured indoor environments, spatial graphs have shown promise in representing room layouts or object co-occurrences \cite{graph_generation}. Our work extends this idea to static environments, explicitly modelling spatial displacements, angles, overlap and object sizes as edge features for downstream anomaly detection and correction of object detection predictions.

In this paper, we propose a graph-based framework for modelling the spatial layout of static environments, such as that of a maritime simulator using egocentric video. By treating each annotated object as a node in a spatial graph, we learn the structural relationships between them and use this information for two downstream tasks: anomaly detection, which identifies mislabeled objects in new scenes, and anomaly correction, which proposes corrected labels based on spatial context. Our approach is evaluated on a dataset collected in a maritime bridge simulator, where egocentric views from a moving participant were manually annotated to create spatially rich scene representations. We simulate annotation errors to train and evaluate a multi-task Graph Neural Network (GNN) that jointly classifies node validity and predicts corrected labels. Additionally, we demonstrate that applying the anomaly correction as a postprocessing step significantly improves the performance of state-of-the-art object detectors such as YOLOv7 and RT-DETR trained on the same dataset.

In summary, our contributions are:
\begin{enumerate}
    \item A spatial graph-based representation of egocentric maritime simulator scenes, capturing both semantic and geometric object relationships.
    \item A multi-task GNN model for node-level anomaly detection and correction, trained using synthetically corrupted scenes.
    \item An empirical analysis of how structural correction improves object detection accuracy in noisy but static environments.
\end{enumerate}

\section{Dataset}
The dataset was collected at the Advanced Navigation Research Simulator (ANRS), located at the Centre of Excellence in Maritime Safety (CEMS), Singapore. Data acquisition was conducted using the scene camera of Tobii Pro Glasses 3, worn by a participant performing standardised navigational tasks. This setup enabled the collection of egocentric video sequences as the participant moved within the simulator environment, offering varied perspectives of a static but spatially complex ship bridge layout. The total number of frames extracted from the videos were 26k.

To construct a spatially-aware visual representation of the simulator, key panels and equipment were manually annotated in each frame using bounding boxes and semantic labels for 39 distinct classes representing equipment and panels in the ship bridge simulator. Each annotated object includes its label, bounding box coordinates, geometric centre, and spatial size. These annotations serve as the basis for building a graph representation of the environment.

\begin{figure}
     \centering
     \begin{subfigure}{0.49\textwidth}
         \centering
         \includegraphics[width=\textwidth, height=0.6 \textwidth]{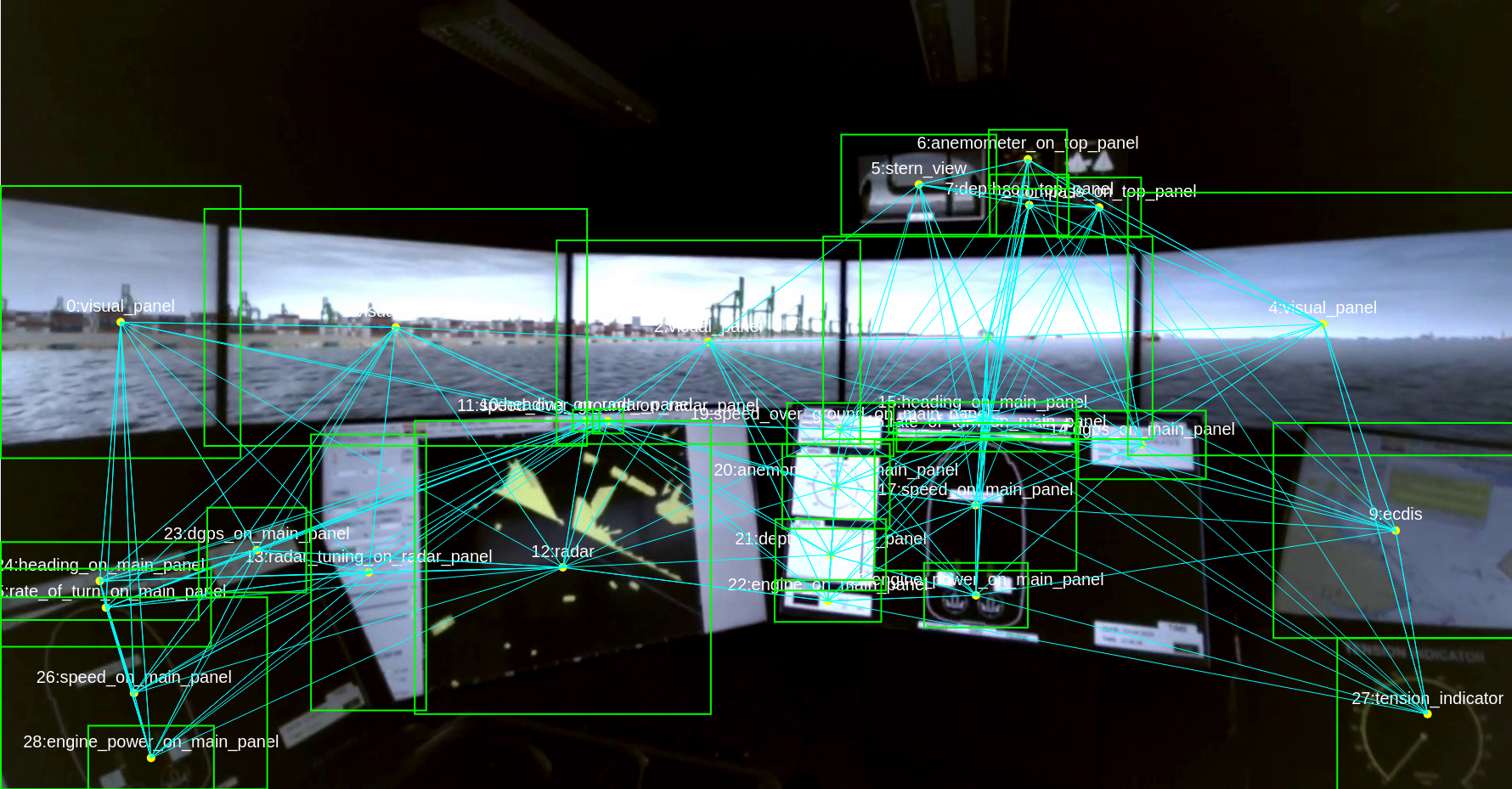}
         \caption{Centre perspective}
     \end{subfigure}
     \begin{subfigure}{0.49\textwidth}
         \centering
         \includegraphics[width=\textwidth, height=0.6 \textwidth]{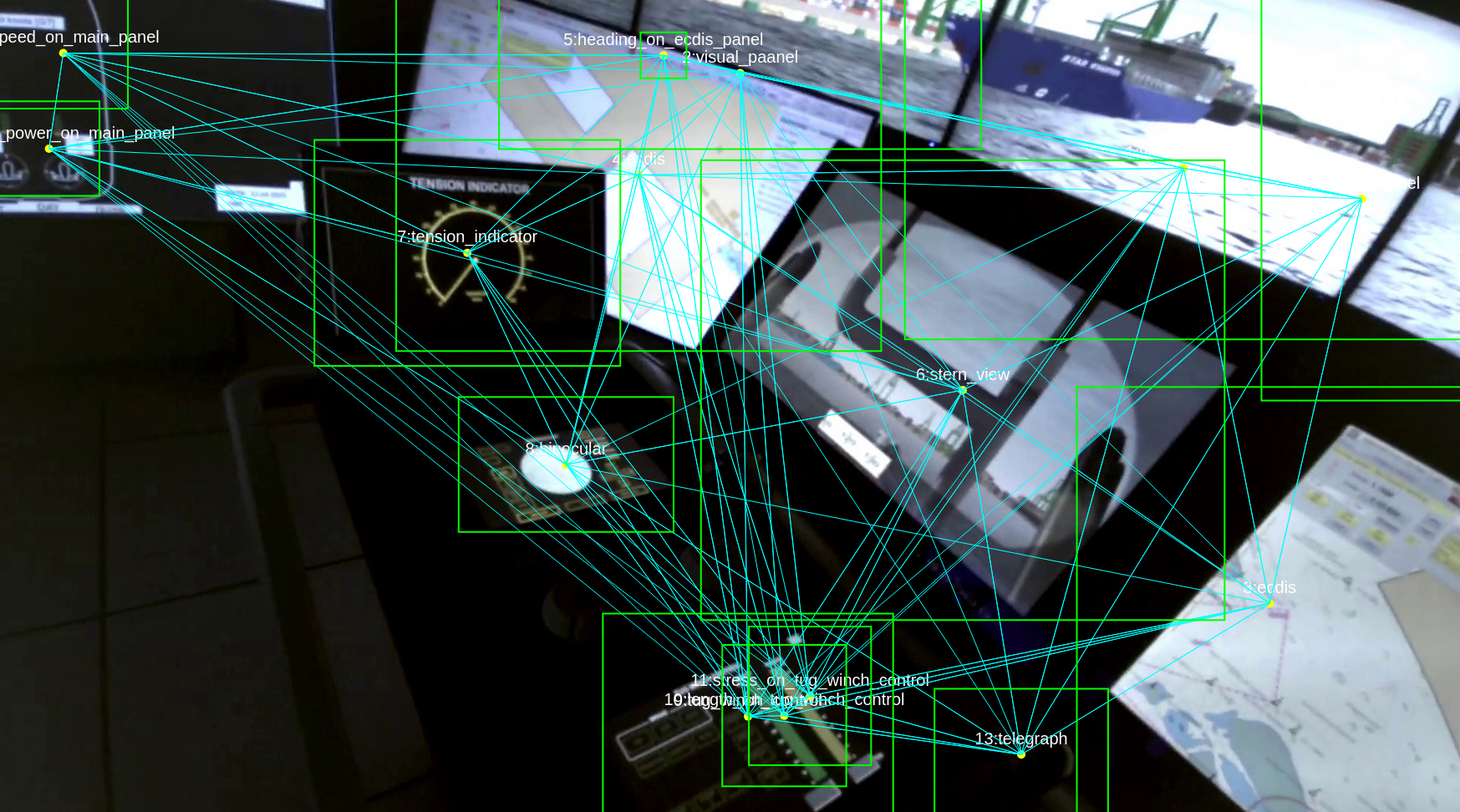}
         \caption{Left perspective}
     \end{subfigure}
     \vfill
     \begin{subfigure}{0.49\textwidth}
         \centering
         \includegraphics[width=\textwidth, height=0.6 \textwidth]{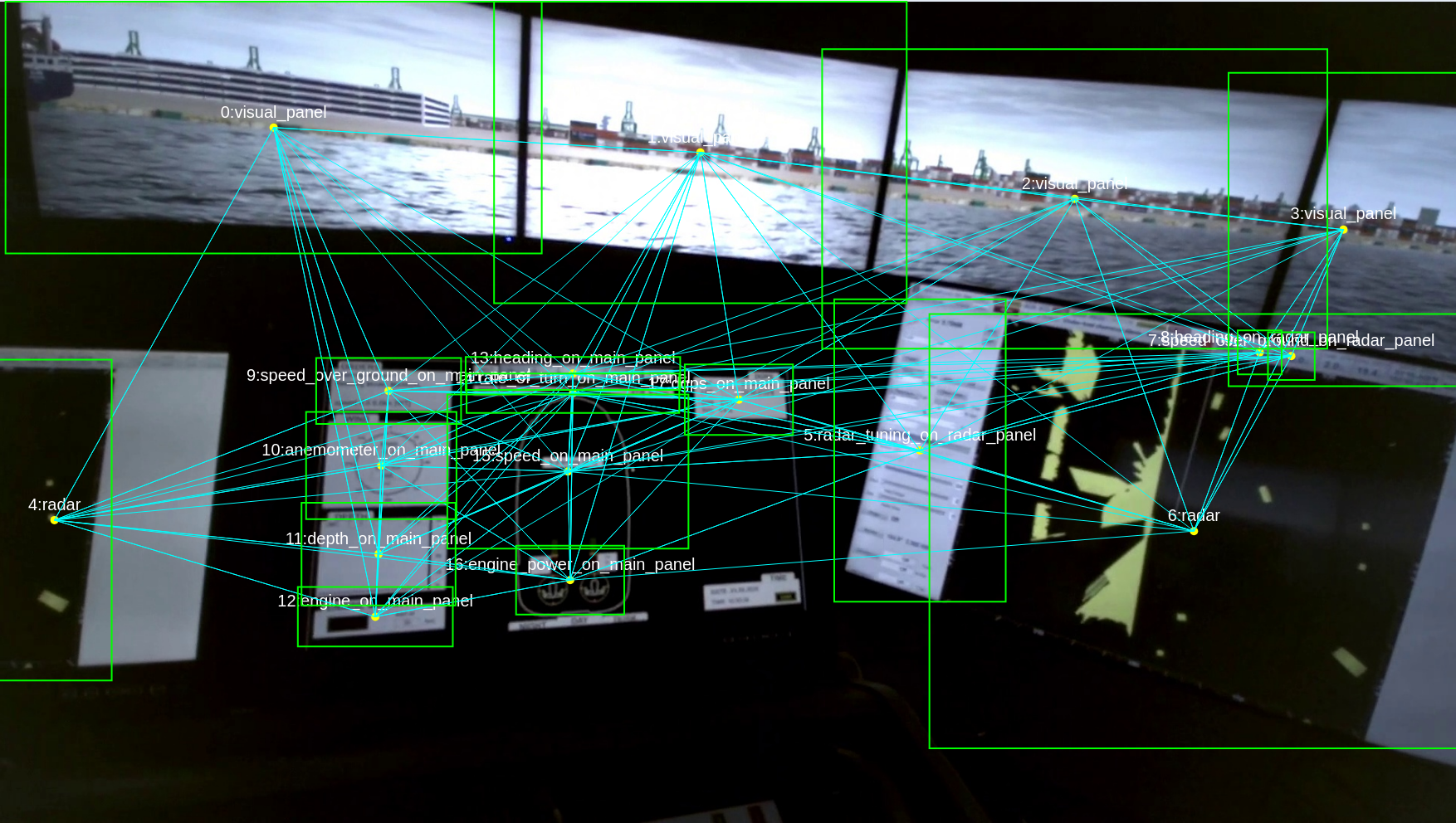}
         \caption{Right perspective}
     \end{subfigure}
     \begin{subfigure}{0.49\textwidth}
         \centering
         \includegraphics[width=\textwidth, height=0.6 \textwidth]{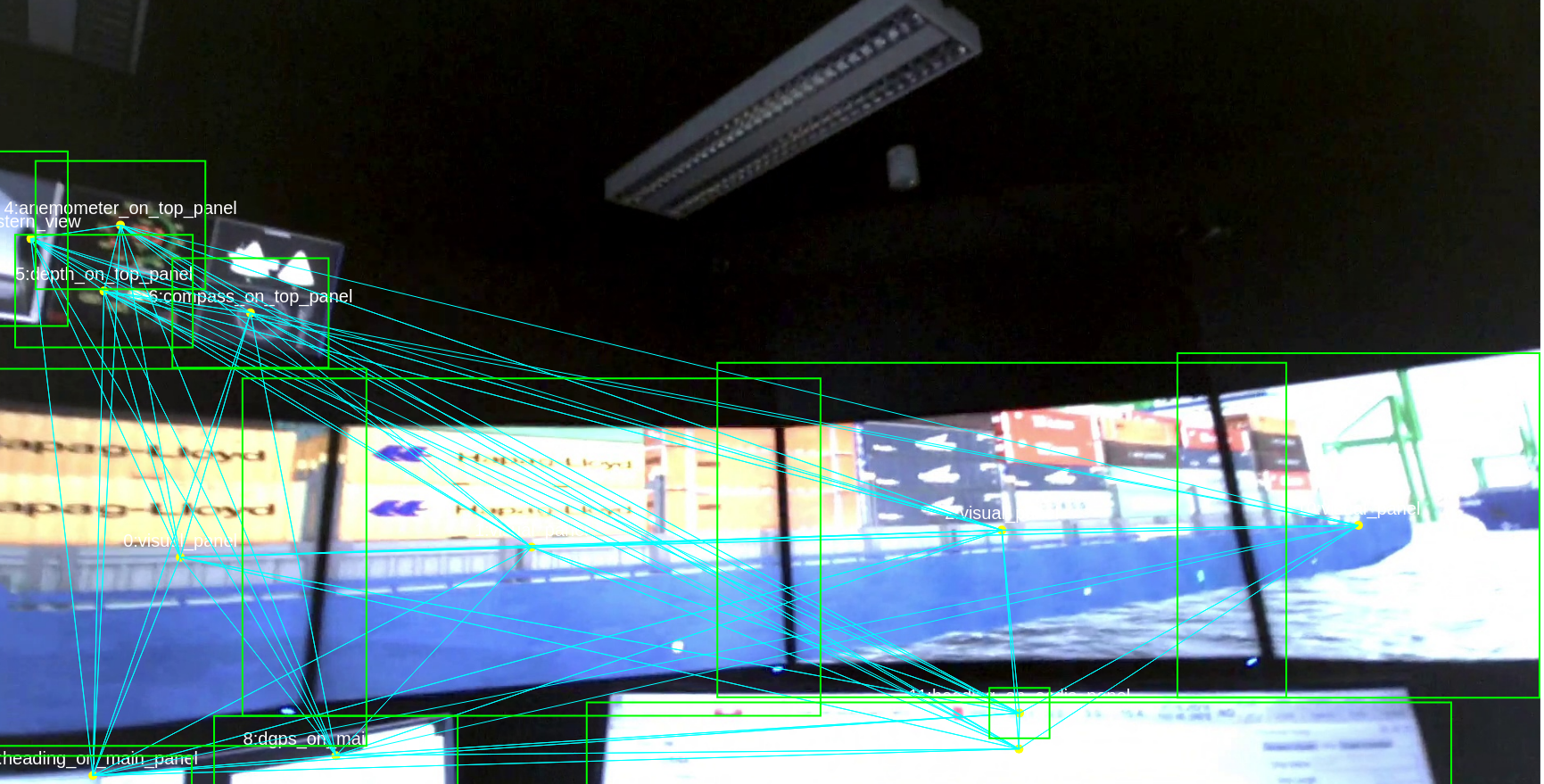}
         \caption{Top right perspective}
     \end{subfigure}
    \caption{Samples from the dataset showing annotated graph nodes with green bounding boxes, node labels in white text, and spatial edges to neighbouring nodes highlighted in cyan.}
    \label{fig:dataset samples}
\end{figure}

For each annotated frame, a graph is constructed where nodes correspond to objects, and node features encode the object’s label, spatial centre, and size. Edges are formed by connecting each node to its $k$ nearest neighbours based on the Euclidean distance between object centres. Edge features capture pairwise spatial relations, including relative displacement, Euclidean distance, and angular orientation, Intersection over Union (IoU) of bounding boxes, and the relative size ratio between objects. \Cref{fig:dataset samples} presents examples of the generated graphs overlaid on egocentric images, illustrating the spatial relationships between annotated panels and equipment from the participant's visual perspective.

Each node $v_i$ represents an object (panel/equipment) in the simulator, with the following feature vector:
\begin{equation}
    \mathbf{f_i} = [label_i, x_i, y_i, w_i, h_i]
    \label{eq:node features}
\end{equation}
where $label_i$ is the object's semantic label, $(x_i, y_i)$ are the coordinates of the object's centre, $(w_i, h_i)$ are the width and height of the object's bounding box.

For each edge $e_{ij}$ between nodes $v_i$ and $v_j$, the feature vector is:
\begin{equation}
    \mathbf{e_{ij}} = [\Delta x_{ij}, \Delta y_{ij}, d_{ij}, \theta _{ij}, IoU_{ij}, \rho_{ij}]
    \label{eq:edge features}
\end{equation}
where, 

$(\Delta x_{ij}, \Delta y_{ij})$ is the horizontal $x_j-x_i$ and vertical $y_j-y_i$ displacement respectively, 

$d_{ij}$ is the Euclidean distance $\sqrt{(x_j-x_i)^2+(y_j-y_i)^2}$, 

$\theta _{ij}$ is the angle in degrees between $v_i$ and $v_j$ measured as $arctan2(y_j-y_i, x_j-x_i) \frac{180}{\pi}$, 

$IoU_{ij}$ is the Intersection over Union of the bounding boxes for $v_i$ and $v_j$ measured as $\frac{A_i \cap A_j}{A_i \cup A_j}$ where $A_i, A_j$ are the areas of $v_i$ and $v_j$ respectively, and 

$\rho_{ij}$ is the relative size ratio between $v_i$ and $v_j$ measured as $\frac{w_jh_j}{w_ih_i}$.

In addition to supporting graph construction, the annotated dataset is also used to train object detection models, which serve as a validation mechanism for the anomaly detection and correction algorithm. This enables a systematic evaluation of the improvement of object detection accuracy using graph-based anomaly corrections.

\section{Proposed Methodology}

\subsection{Data Augmentation} 
To train and evaluate the robustness of anomaly detection and correction models, we augment the original dataset with synthetic negative samples. These modified samples simulate realistic deviations from the expected spatial configuration. Specifically:
\begin{enumerate}
    \item A random subset of objects (up to $\rho$ per frame) are selected.
    \item Their labels are replaced with incorrect alternatives sampled from the full set of valid equipment labels for semantic misalignment. 
    \item Positional jitter is applied to object bounding boxes by adding Gaussian noise to the corner coordinates, simulating annotation drift or perception error.
    \item Nodes corresponding to modified annotations are labelled as invalid, while untouched nodes are labelled as valid.
\end{enumerate}

\subsection{Model Architecture}
To address the tasks of anomaly detection and anomaly correction at the node level, we design a graph-based multi-task model built upon the GraphSAGE architecture. The model processes spatial object graphs constructed from egocentric scene views and outputs both validity scores and label predictions for each node. 

\subsubsection{Graph Encoder}
Let $G=(V,E)$ be an undirected graph where each node $v_i \in V$ corresponds to an object class in a static environment. Each node is associated with a feature vector $\mathbf{f_i} \in R^5$ encoding the object's features as detailed in \cref{eq:node features}.  Each edge $e_{ij} \in E$ corresponds to the spatial relationships between objects and is represented using the edge features as described in \cref{eq:edge features}. The graph encoder consists of two stacked GraphSAGE convolution layers with a mean aggregator, which learn node embeddings by aggregating information from neighbours. 
\begin{equation}
    \mathbf{h_i}^{(1)} = ReLU \left( SAGEConv^1(\mathbf{f_i},\mathbf{f_j}|j\in \mathcal{N}(i)\right)
\end{equation}    
\begin{equation}
    \mathbf{h_i}^{(2)} = ReLU \left( SAGEConv^2(\mathbf{h_i}^{(1)},\mathbf{h_j}^{(1)}|j\in \mathcal{N}(i)\right)
\end{equation}
where $\mathcal{N}(i)$ represents the neighbours of node $v_i$. Here $\mathbf{h_i}^{(2)} \in R^{64}$ denotes the final embedding of node $v_i$ where 64 is the number of hidden features.

\subsubsection{Multi-Task heads}
The final node embeddings are passed through two task-specific output heads. The anomaly detection head is a linear layer that maps each node embedding to a scalar logit, which is used to predict whether the node is valid (i.e., unmodified) or anomalous.
\begin{equation}
    \hat{v_i} = \sigma(\mathbf{W_v}^{valid} \mathbf{h_i}^{(2)} +b_v^{valid})
\end{equation}
where $\sigma$ is the sigmoid activation and $\hat{v_i} \in [0,1]$ indicates the predicted validity for node $v_i$. $W_v^{valid}$ and $b_v^{valid}$ are the learned weights and biases, respectively.

The anomaly correction head is a separate linear layer that projects the node embedding into an $n$-dimensional space ($n$ is the number of unique classes), producing logits for a softmax-based label prediction.
\begin{equation}
    \hat{y_i} = softmax(\mathbf{W_y}^{label} \mathbf{h_i}^{(2)} +\mathbf{b_y}^{label})
\end{equation}
where $\hat{y} \in R^n$ represents the predicted class probabilities for node $v_i$. $W_v^{label}$ and $b_v^{label}$ are the learned weights and biases, respectively.

\subsubsection{Training Objective}
The model is jointly trained end-to-end using a multi-task loss combining binary cross-entropy for validity prediction and categorical cross-entropy for label classification. 
\begin{equation}
    \mathcal{L} = \lambda_{valid}\mathcal{L}_{BCE}(\hat{v_i},val_i) +\lambda_{label}\mathcal{L}_{CE}(\hat{y_i},y_i) 
\end{equation}
where $val_i, y_i$ are the ground truth labels for validity and node class label for node $v_i$ respectively, and $\lambda_{valid}, \lambda_{label}$ are the weighting coefficients.

\subsubsection{Training}
The model is trained using mini-batch stochastic gradient descent with the Adam optimiser and a learning rate of $0.001$. Training spans 30 epochs on the spatially constructed graph samples, including both clean and synthetically corrupted samples. The train, validation, and test samples are split in a 70:15:15. The model is optimised using the multi-task training objective, and training is performed on NVIDIA GeForce RTX 4070. 

\subsection{Experiment}
We design a series of experiments to evaluate the proposed graph-based anomaly detection and correction framework and its impact on downstream object recognition tasks. To study the impact of graph construction and anomaly simulation parameters, we conduct an ablation study to examine how two key parameters influence the performance of the model: the neighbourhood size $k$ which controls the number of edges for each node, and anomaly ratio $\rho$ which denotes the maximum number of object labels modified in a single negative sample. For each configuration, we measure the performance of the model on a fixed test set. We assess the practical utility of the anomaly correction model by studying its effect on standard object detection models trained on the same dataset. Specifically, we compare the performance of two state-of-the-art detectors, YOLOv7 and RT-DETR. Each model is trained on the original annotated dataset and evaluated on a common test set. The GNN-based anomaly detection and correction module was applied to the raw detection outputs, treating them as graph nodes and refining their predictions. This setup allows us to quantify how much performance recovery is enabled by the graph-based correction branch. 

\subsection{Evaluation Metrics}
We evaluate the proposed model using two distinct sets of metrics. The node-level classifier is evaluated using the following metrics:
\begin{enumerate}
    \item Validity Accuracy measures the ability of the model to distinguish between valid and invalid nodes. This is a standard binary classification accuracy represented as $\frac{1}{N}\sum_{i=1}^N1[\hat{v_i}=val_i]$
    \item Label Prediction Scores including accuracy, weighted average precision, recall, and F1 score over classes computed using the ground truth label and the predicted label for nodes classified as invalid. 
    \item Confusion Matrix to visualise model performance across object classes, highlighting common misclassifications.
\end{enumerate}
To assess how much the anomaly correction branch improves end-to-end recognition, we evaluate on the same test set before and after applying the correction model on the object detection model outputs using the standard object detection metric:
\begin{enumerate}
    \item mAP@50 (mean Average Precision at IoU $\geq$ 0.5) to measure average precision across object classes with a 0.5 IoU threshold for a true positive. We report mAP@50 as the primary evaluation metric for object detection, as the proposed anomaly correction model is designed to improve classification accuracy rather than the precision of bounding box localisation. Since our corrections target mislabeled or misclassified objects and do not directly modify bounding box geometry, improvements are expected to manifest more clearly at a lower IoU threshold (0.5), where class correctness dominates the evaluation. Higher IoU thresholds, which emphasise fine-grained localisation, are less sensitive to the semantic corrections that our model provides.
\end{enumerate}

\section{Results}

\subsection{Qualitative Results}
\begin{figure}
     \centering
     \begin{subfigure}{0.49\textwidth}
         \centering
         \includegraphics[width=\textwidth, height=0.6 \textwidth]{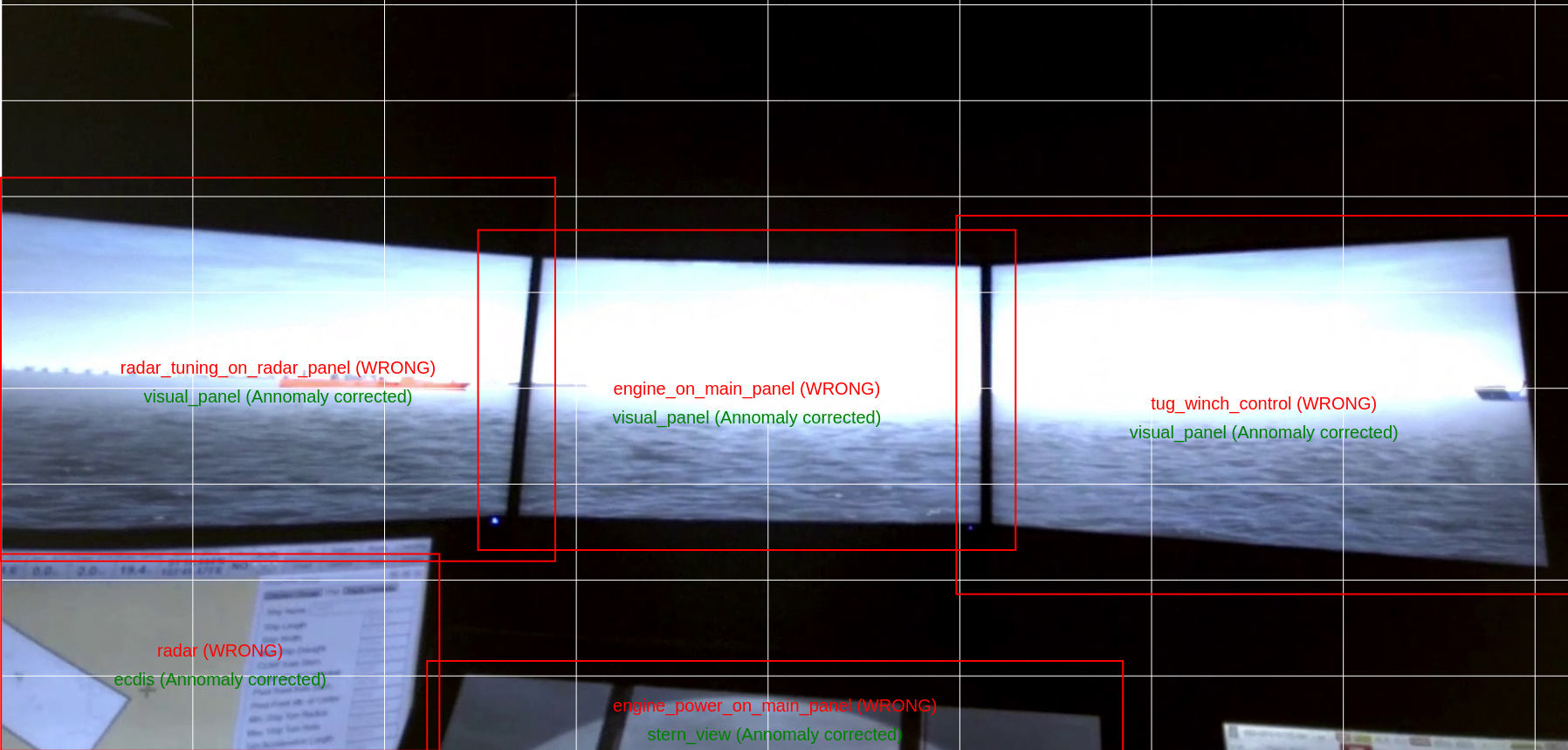}
         \caption{}
     \end{subfigure}
     \begin{subfigure}{0.49\textwidth}
         \centering
         \includegraphics[width=\textwidth, height=0.6 \textwidth]{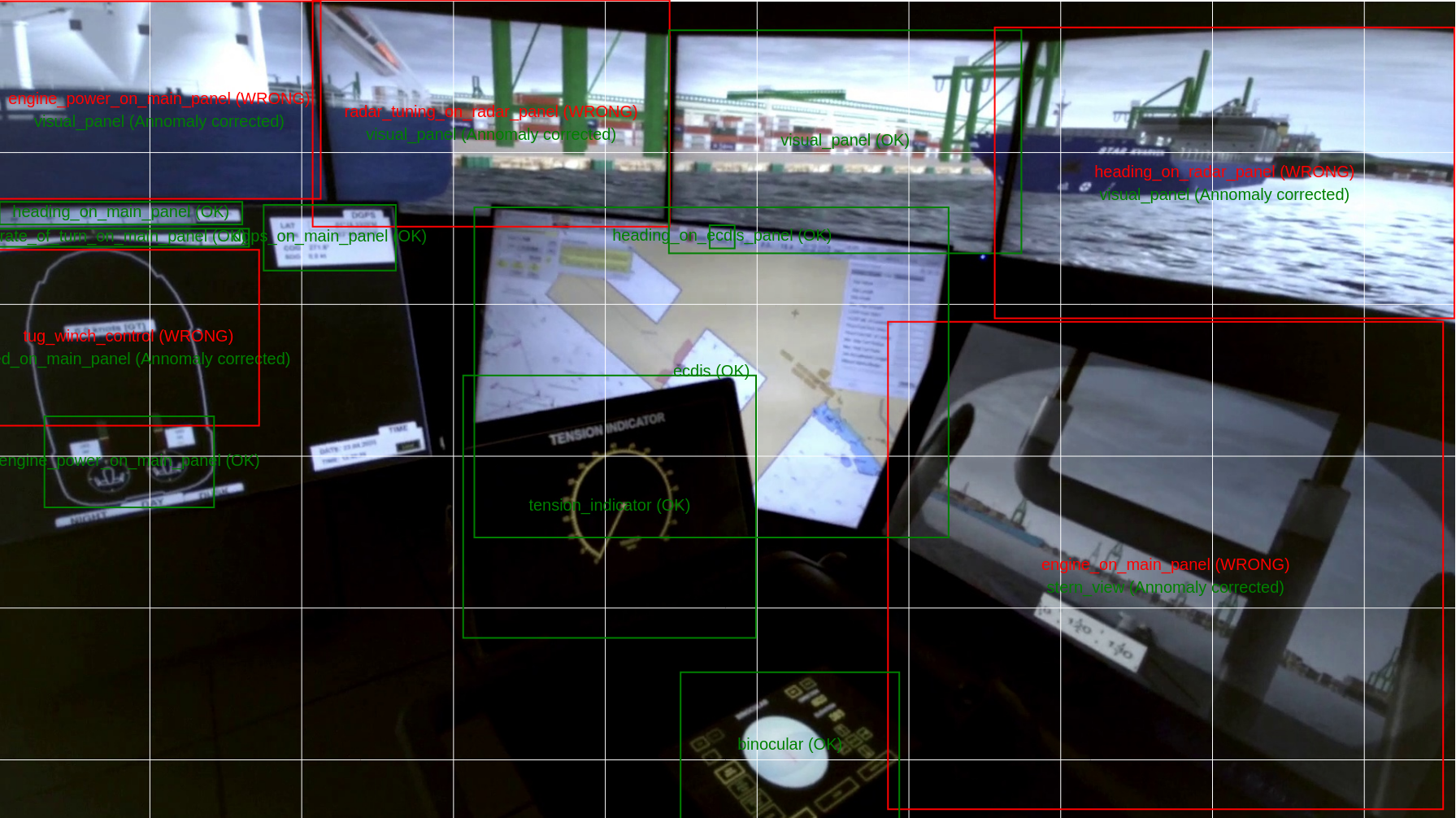}
         \caption{}
     \end{subfigure}
    \caption{Randomly selected negative samples from the test dataset showing results of anomaly detection (bounding boxes in red) and anomaly correction (corresponding corrected label in green)}
    \label{fig:test samples}
\end{figure}

\Cref{fig:test samples} illustrates qualitative results on randomly sampled negative samples from the test set, where the proposed model successfully identifies anomalous annotations and suggests corrected labels. These examples demonstrate the effectiveness of the anomaly detection and correction pipeline in recovering semantically consistent object annotations.

\subsection{Ablation Study Results}
\begin{figure*}
     \centering
     \begin{subfigure}{0.24\textwidth}
         \centering
         \includegraphics[width=\textwidth]{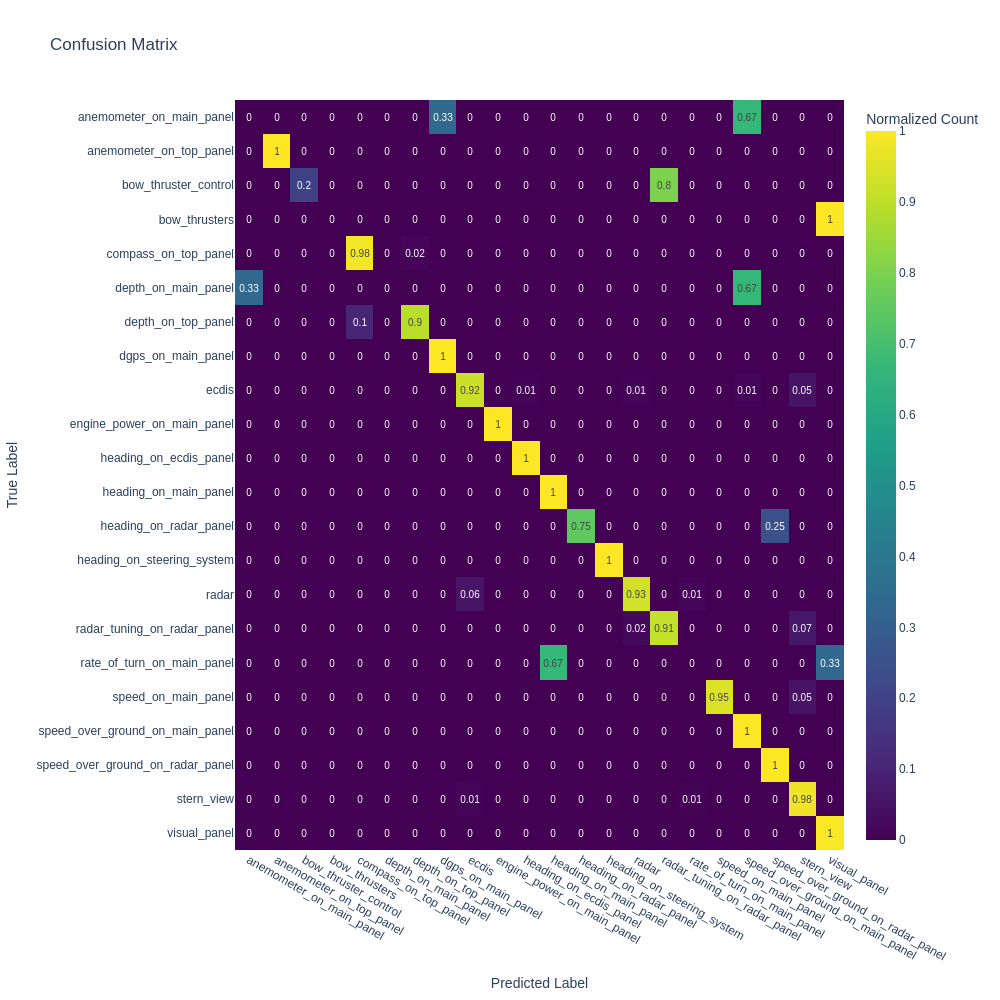}
         \caption{$\rho =  1, k=5$}
     \end{subfigure}
     \begin{subfigure}{0.24\textwidth}
         \centering
         \includegraphics[width=\textwidth]{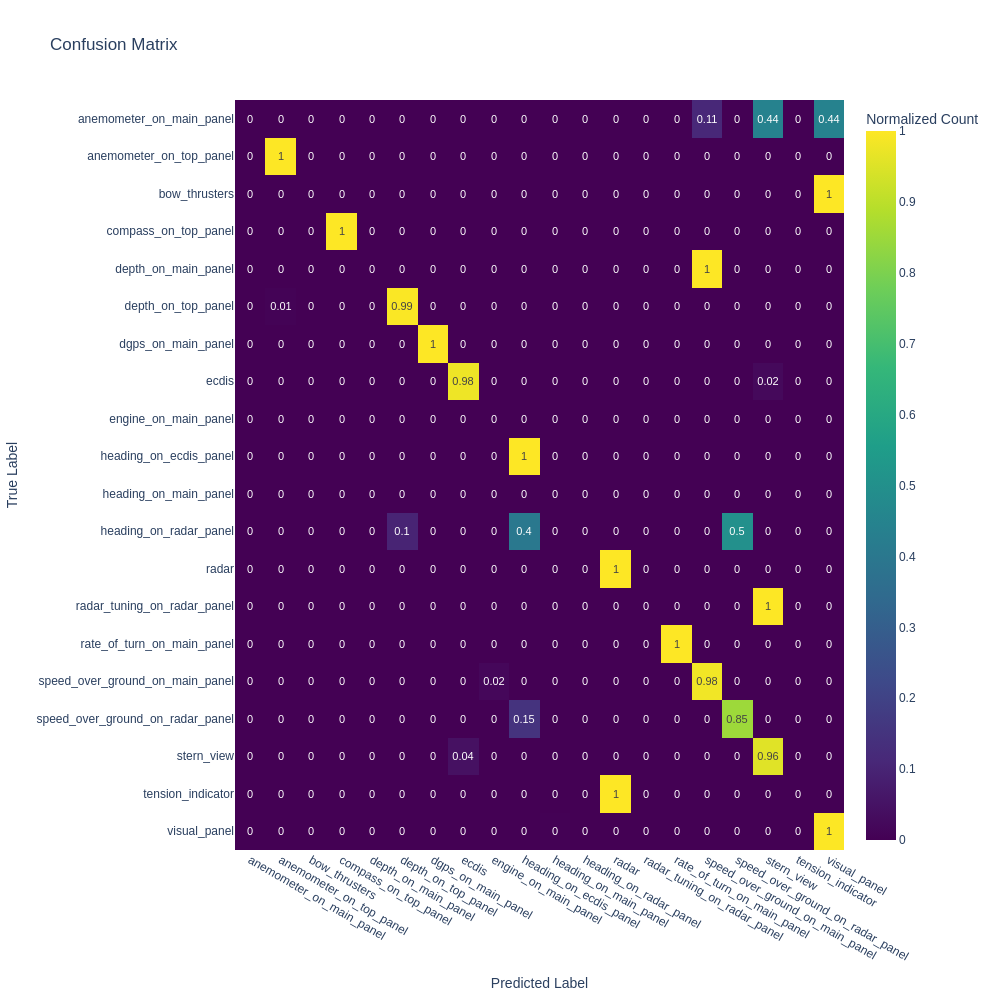}
         \caption{$\rho =  1, k=7$}
     \end{subfigure}
     \begin{subfigure}{0.24\textwidth}
         \centering
         \includegraphics[width=\textwidth]{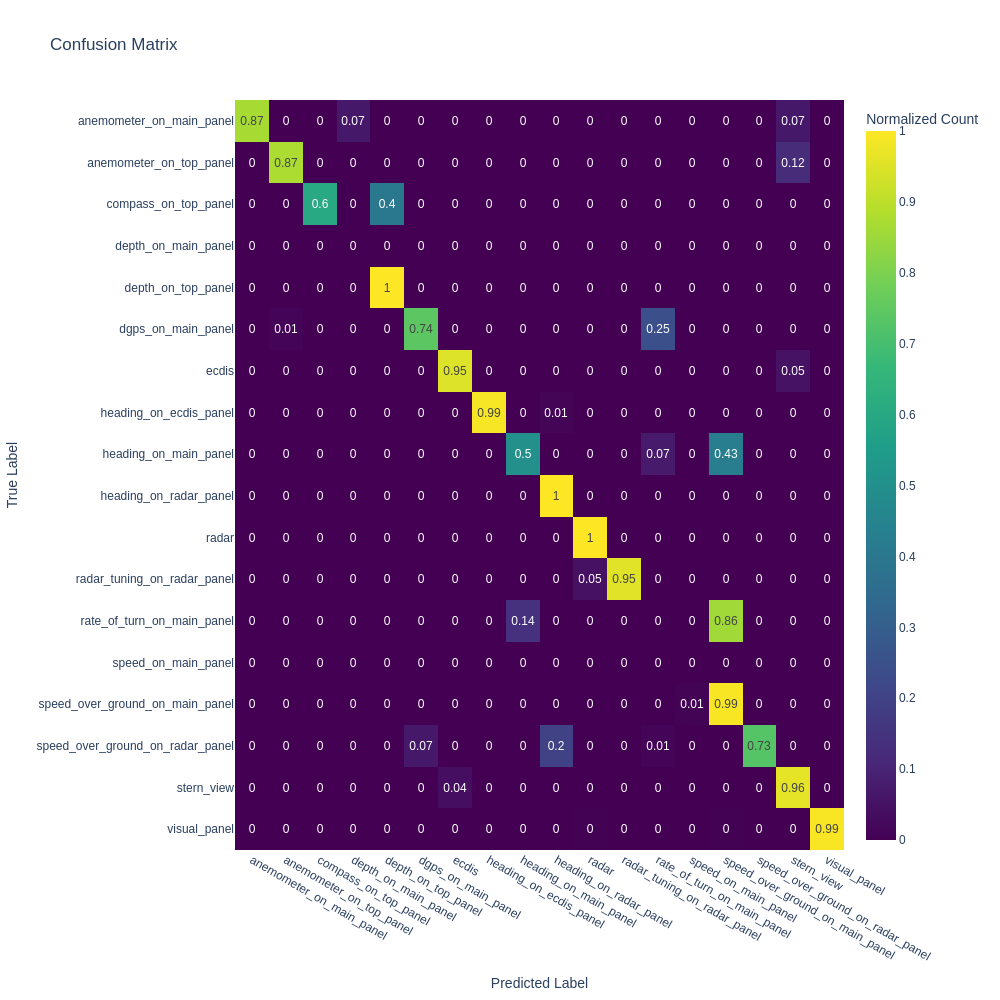}
         \caption{$\rho =  1, k=10$}
     \end{subfigure}
     \begin{subfigure}{0.24\textwidth}
         \centering
         \includegraphics[width=\textwidth]{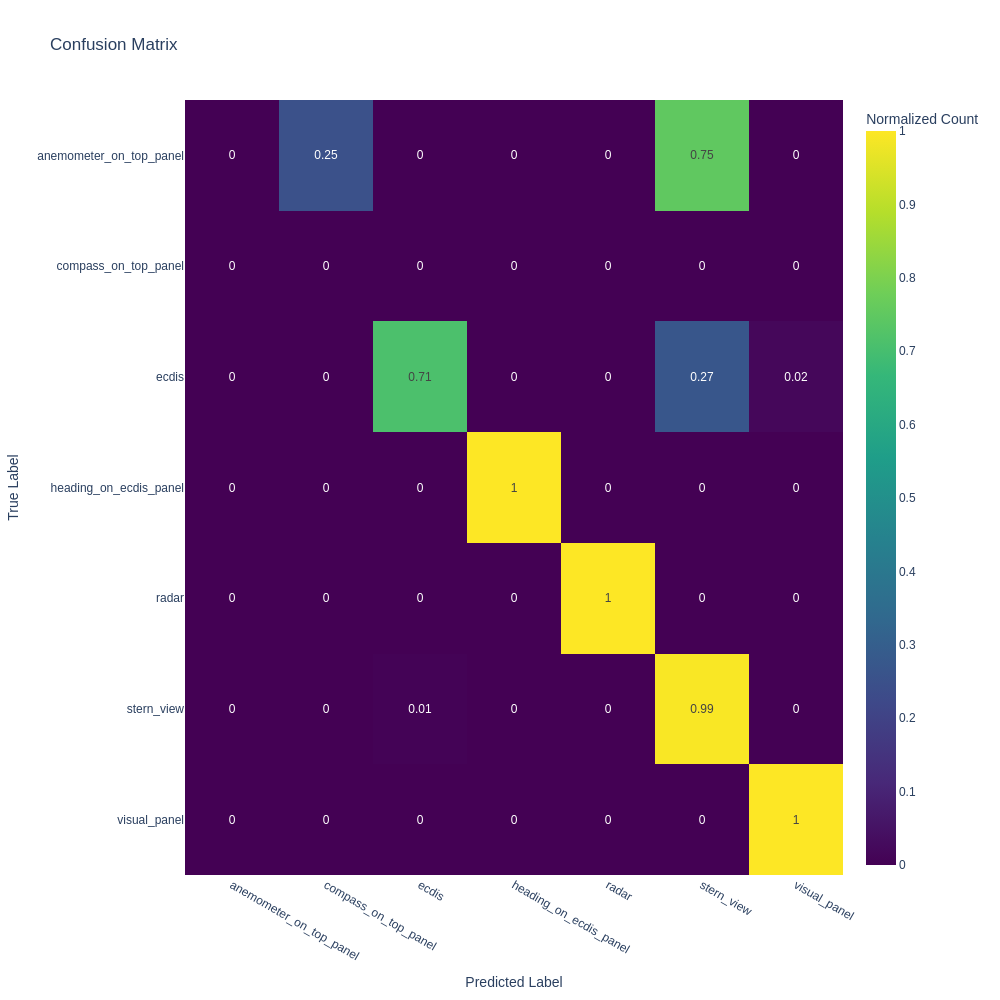}
         \caption{$\rho =  1, $ all neighbours}
     \end{subfigure}
     \vfill
     \begin{subfigure}{0.24\textwidth}
         \centering
         \includegraphics[width=\textwidth]{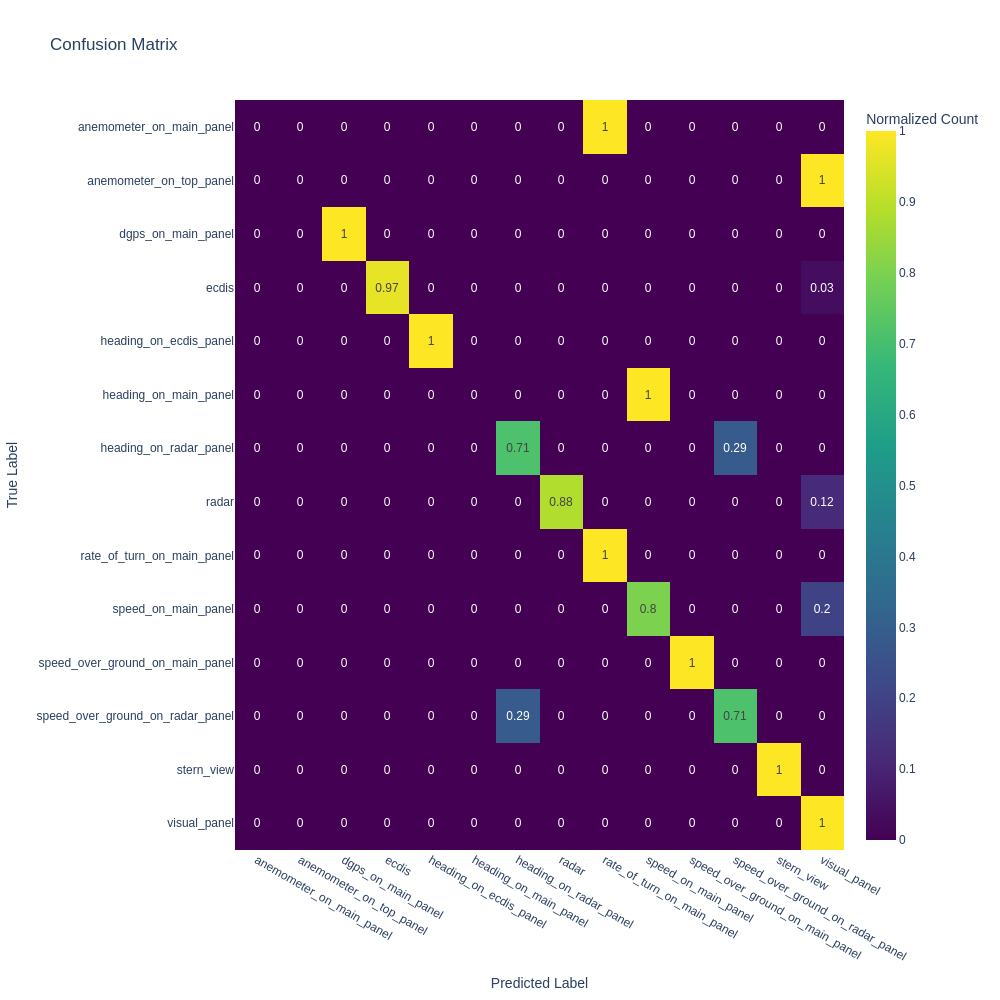}
         \caption{$\rho =  3, k=5$}
     \end{subfigure}
     \begin{subfigure}{0.24\textwidth}
         \centering
         \includegraphics[width=\textwidth]{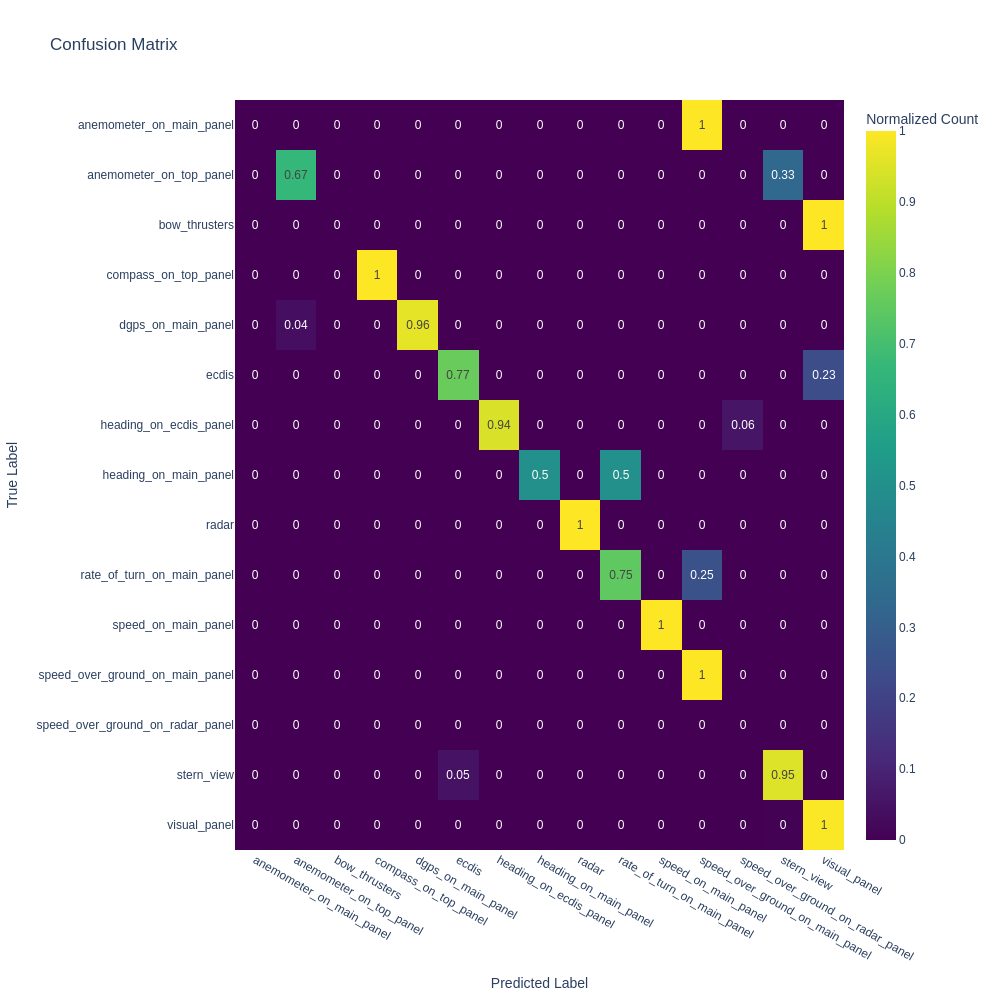}
         \caption{$\rho =  3, k=7$}
     \end{subfigure}
     \begin{subfigure}{0.24\textwidth}
         \centering
         \includegraphics[width=\textwidth]{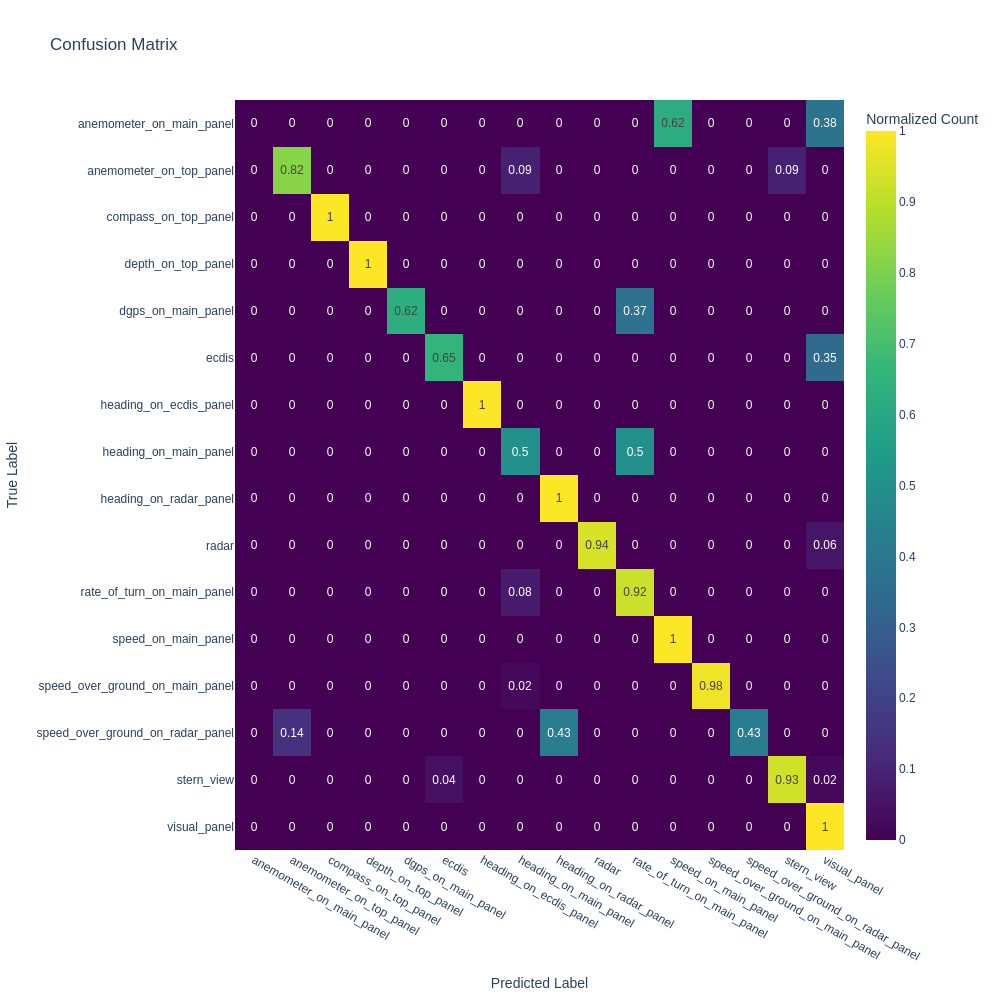}
         \caption{$\rho =  3, k=10$}
     \end{subfigure}
     \begin{subfigure}{0.24\textwidth}
         \centering
         \includegraphics[width=\textwidth]{confusion_matrix_k_7_p_1.png}
         \caption{$\rho =  3, $ all neighbours}
     \end{subfigure}
     \vfill
     \begin{subfigure}{0.24\textwidth}
         \centering
         \includegraphics[width=\textwidth]{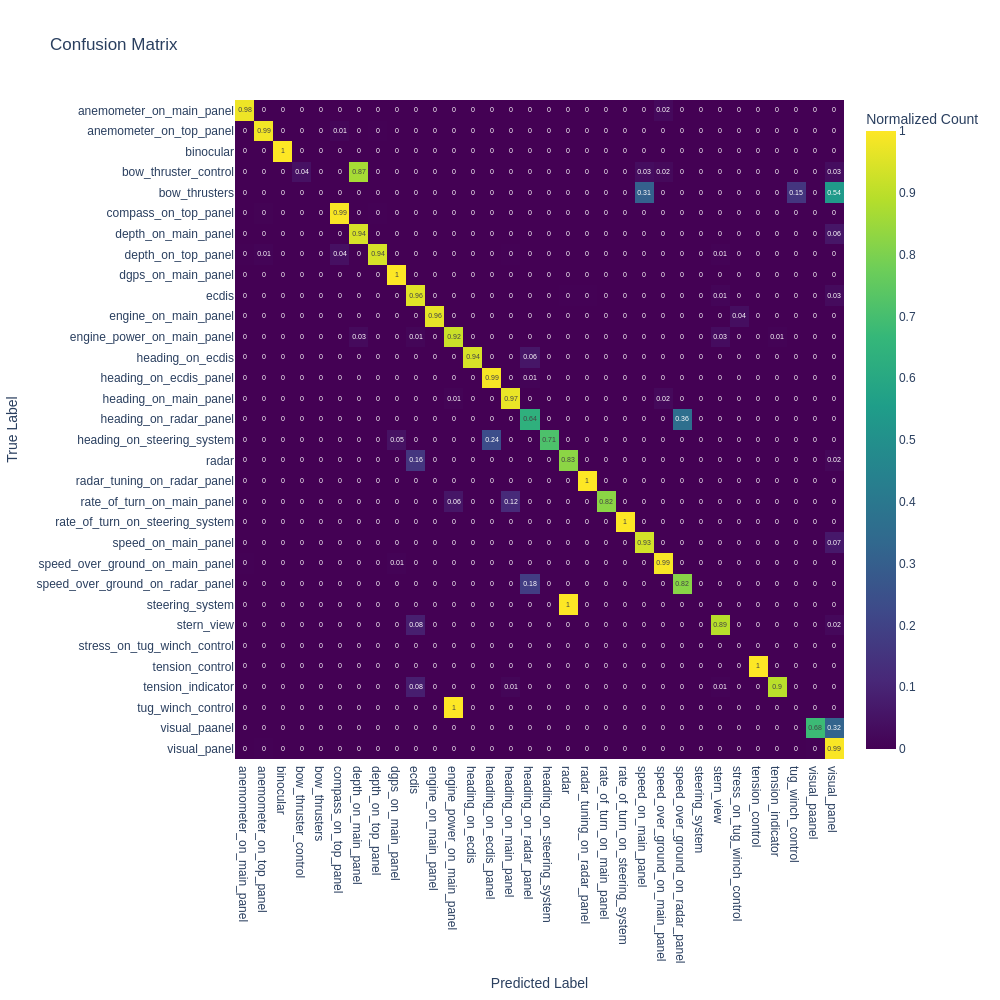}
         \caption{$\rho =  5, k=5$}
     \end{subfigure}
     \begin{subfigure}{0.24\textwidth}
         \centering
         \includegraphics[width=\textwidth]{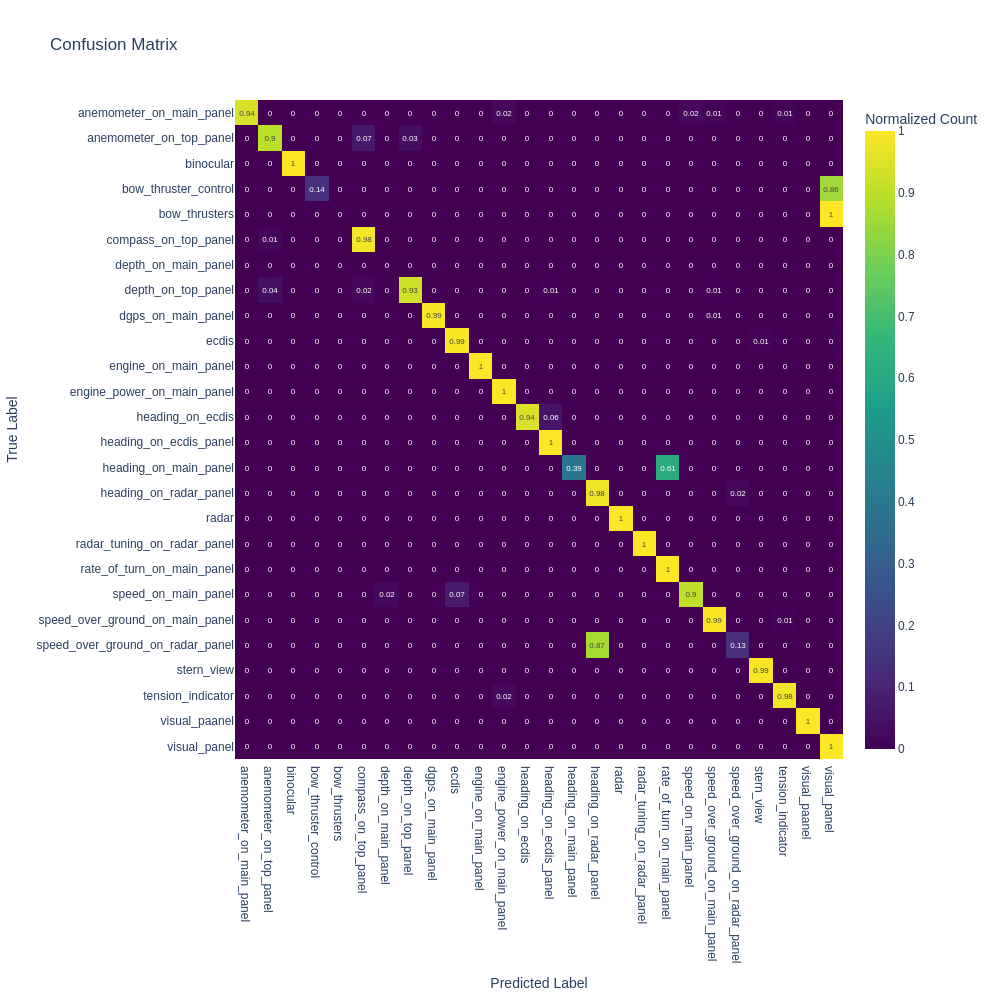}
         \caption{$\rho =  5, k=7$}
     \end{subfigure}
     \begin{subfigure}{0.24\textwidth}
         \centering
         \includegraphics[width=\textwidth]{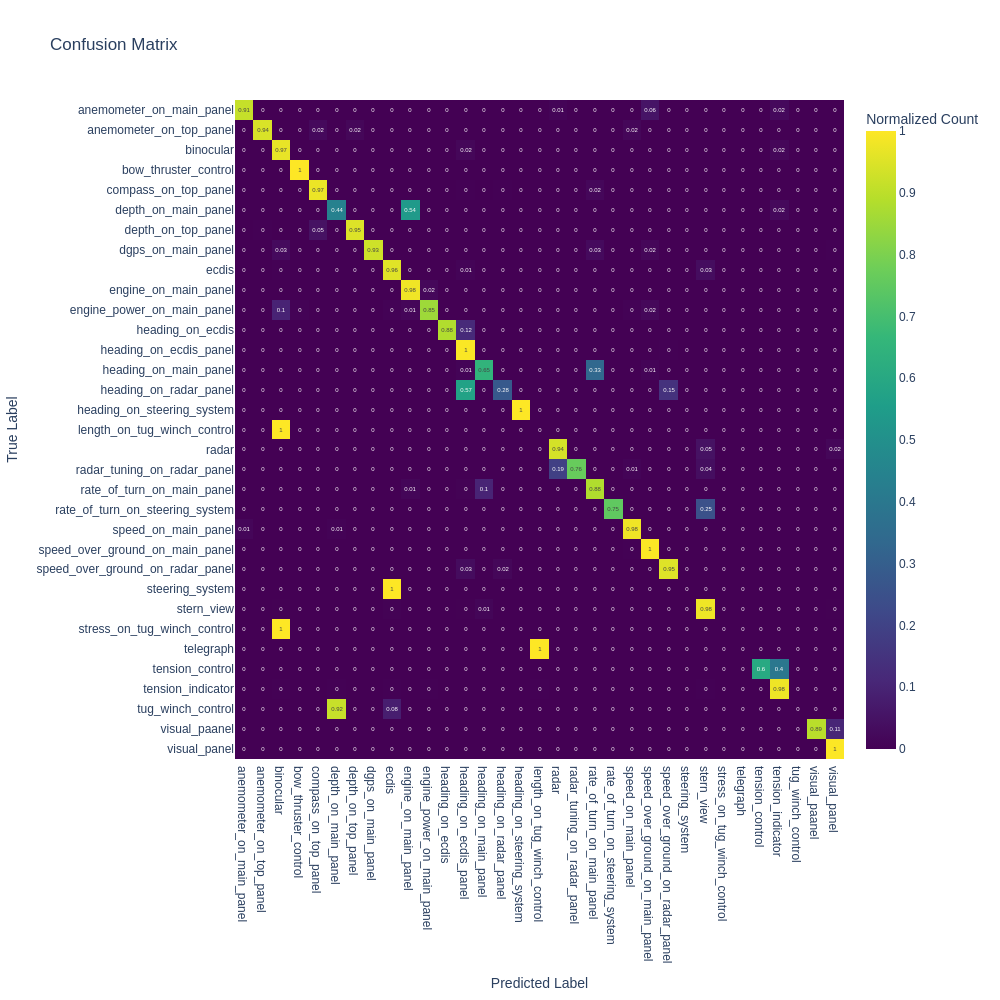}
         \caption{$\rho =  5, k=10$}
     \end{subfigure}
     \begin{subfigure}{0.24\textwidth}
         \centering
         \includegraphics[width=\textwidth]{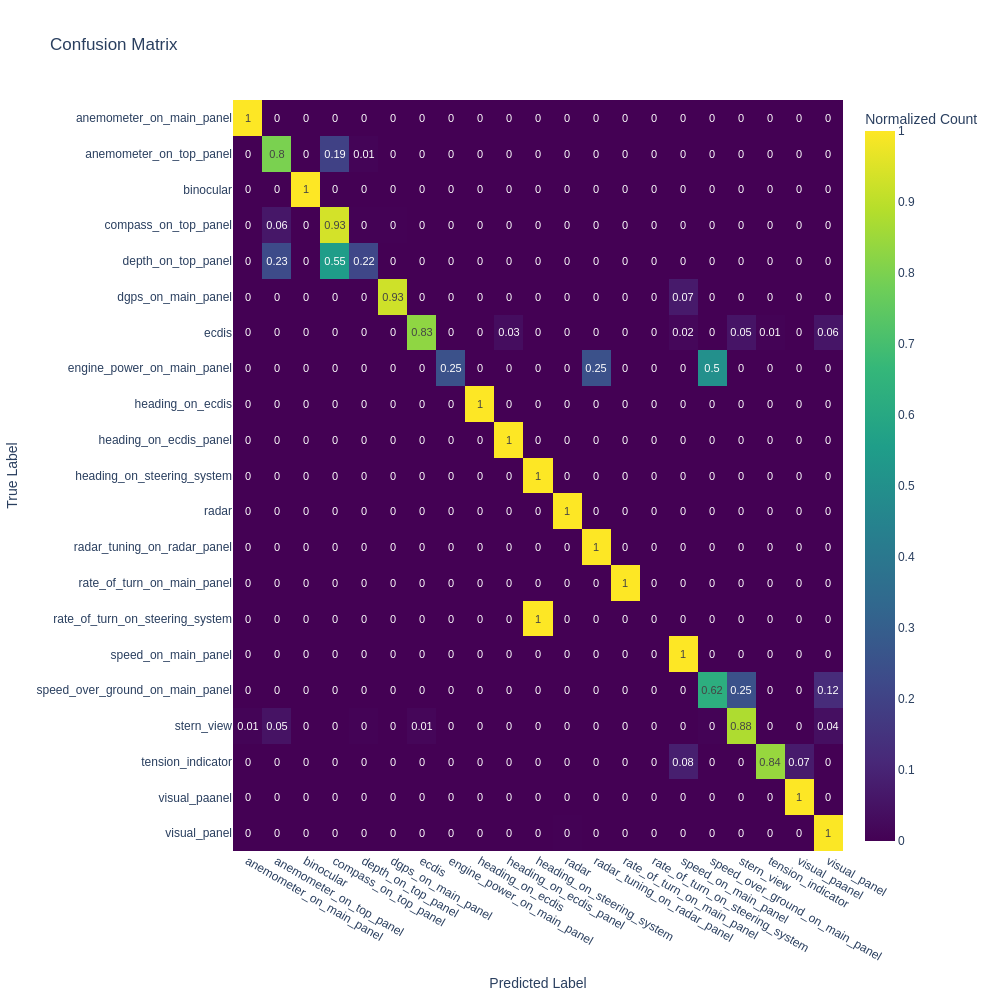}
         \caption{$\rho =  5, $ all neighbours}
     \end{subfigure}
    \caption{Confusion matrices generated from various ablation study experiments. In some settings with lower values of $\rho$, certain classes were absent in the test set due to the nature of negative sample generation, and thus are not represented in the corresponding matrices.}
    \label{fig:confusion matrix}
\end{figure*}

\Cref{tab:ablation results} presents the validity classification accuracy and the label correction performance (accuracy, weighted precision, recall, and F1-score) across various combinations of $k$, the maximum number of neighbours, and $\rho$, the number of objects that are randomly modified in each generated negative sample. \Cref{fig:confusion matrix} shows the corresponding confusion matrices for the experiments.

Across values of $\rho$, using a moderate neighbourhood size ($k=5$ or $k = 7$) yields optimal or near-optimal performance for anomaly detection as demonstrated in the validity accuracy. For smaller anomaly ratios ($\rho = 1$), the model is relatively robust across $k$, achieving high validity accuracy (above 0.98) and F1-scores above 0.97. However, with higher anomaly ratios ($\rho =3$ and $\rho = 5$), performance tends to drop for smaller or excessively large $k$, likely due to under- or over-smoothing. Notably, using all neighbours leads to consistently lower label prediction F1-scores, indicating that unrestricted context aggregation can introduce noise.

As expected, increasing the number of perturbed objects per negative sample makes the correction task more challenging. For instance, at $k=10$, the validity accuracy decreases from 0.9831 ($\rho = 1$) to 0.9550 ($\rho = 5$), and F1-score drops from 0.977 to 0.938. Despite this degradation, the model remains reasonably robust even under higher corruption levels, maintaining F1-scores above 0.93 in most settings.

\subsection{Impact of GNN-based Post-processing on Object Detection Performance}
\Cref{tab:postprocessing impact} summarises the improvement in detection performance, as measured by mean Average Precision at IoU threshold 0.5 (mAP@50), on the test dataset using state-of-the-art object detection models. Both detectors show a consistent and notable increase in performance after applying GNN-based correction.

\begin{table}[]
\centering
\resizebox{0.9\textwidth}{!}{
\begin{tabular}{|lll|}
\hline
\multicolumn{1}{|l|}{\textbf{\begin{tabular}[c]{@{}l@{}}Neighbourhood\\ Size\end{tabular} }} & \multicolumn{1}{l|}{\textbf{\begin{tabular}[c]{@{}l@{}}Validity Accuracy\\ on test dataset\end{tabular}}} & \textbf{\begin{tabular}[c]{@{}l@{}}Label Prediction Scores\\ on test dataset\end{tabular}} \\ \hline
\multicolumn{3}{|c|}{$\rho$ = 1 (upto 1 object modified in each negative sample)} \\ \hline
\multicolumn{1}{|l|}{k=5} & \multicolumn{1}{l|}{0.9847} & \begin{tabular}[c]{@{}l@{}}A = 0.973\\ P = 0.971\\ R = 0.973\\ F1 = 0.971\end{tabular} \\ \hline
\multicolumn{1}{|l|}{k=7} & \multicolumn{1}{l|}{0.9828} & \begin{tabular}[c]{@{}l@{}}A = 0.979\\ P = 0.968\\ R = 0.979\\ F1 = 0.974\end{tabular} \\ \hline
\multicolumn{1}{|l|}{k=10} & \multicolumn{1}{l|}{0.9831} & \begin{tabular}[c]{@{}l@{}}A = 0.979\\ P = 0.980\\ R = 0.979\\ F1 = 0.977\end{tabular} \\ \hline
\multicolumn{1}{|l|}{all neighbours} & \multicolumn{1}{l|}{0.9732} & \begin{tabular}[c]{@{}l@{}}A = 0.979\\ P = 0.980\\ R = 0.979\\ F1 = 0.977\end{tabular} \\ \hline
\multicolumn{3}{|c|}{$\rho$ = 3 (upto 3 objects modified in each negative sample)} \\ \hline
\multicolumn{1}{|l|}{k=5} & \multicolumn{1}{l|}{0.9757} & \begin{tabular}[c]{@{}l@{}}A = 0.986\\ P = 0.985\\ R = 0.986\\ F1 = 0.985\end{tabular} \\ \hline
\multicolumn{1}{|l|}{k=7} & \multicolumn{1}{l|}{0.9741} & \begin{tabular}[c]{@{}l@{}}A = 0.966\\ P = 0.966\\ R = 0.966\\ F1 = 0.965\end{tabular} \\ \hline
\multicolumn{1}{|l|}{k=10} & \multicolumn{1}{l|}{0.9802} & \begin{tabular}[c]{@{}l@{}}A = 0.944\\ P = 0.946\\ R = 0.944\\ F1 = 0.940\end{tabular} \\ \hline
\multicolumn{1}{|l|}{all neighbours} & \multicolumn{1}{l|}{0.9806} & \begin{tabular}[c]{@{}l@{}}A = 0.856\\ P = 0.855\\ R = 0.856\\ F1 = 0.837\end{tabular} \\ \hline
\multicolumn{3}{|c|}{$\rho$ = 5 (upto 5 objects modified in each negative sample)} \\ \hline
\multicolumn{1}{|l|}{k=5} & \multicolumn{1}{l|}{0.9425} & \begin{tabular}[c]{@{}l@{}}A = 0.95\\ P = 0.956\\ R = 0.950\\ F1 = 0.949\end{tabular} \\ \hline
\multicolumn{1}{|l|}{k=7} & \multicolumn{1}{l|}{0.9498} & \begin{tabular}[c]{@{}l@{}}A = 0.969\\ P = 0.974\\ R = 0.969\\ F1 = 0.963\end{tabular} \\ \hline
\multicolumn{1}{|l|}{k=10} & \multicolumn{1}{l|}{0.9550} & \begin{tabular}[c]{@{}l@{}}A = 0.944\\ P = 0.946\\ R = 0.944\\ F1 = 0.938\end{tabular} \\ \hline
\multicolumn{1}{|l|}{all neighbours} & \multicolumn{1}{l|}{0.9237} & \begin{tabular}[c]{@{}l@{}}A = 0.847\\ P = 0.918\\ R = 0.847\\ F1 = 0.835\end{tabular} \\ \hline
\end{tabular}}
\caption{Validity and label prediction metrics for ablation study over maximum number of neighbours $k$ and anomaly ratio $\rho$ on test dataset (A=Accuracy, P=Weighted Precision over all classes, R=Weighted Recall over all classes, F1= Weighted F1 over all classes)}
\label{tab:ablation results}
\end{table}

\textbf{\begin{table}[]
\begin{tabular}{|l|l|l|}
\hline
\textbf{\begin{tabular}[c]{@{}l@{}}Object Detection \\ Model\end{tabular}} & \textbf{\begin{tabular}[c]{@{}l@{}}mAP50 on \\ test dataset\\ without GNN \\ post-processing\end{tabular}} & \textbf{\begin{tabular}[c]{@{}l@{}}mAP50 on \\ test dataset\\ with GNN \\ post-processing\end{tabular}} \\ \hline
YOLOv7 & 0.891 & 0.927 \\ \hline
RT-DETR & 0.903 & 0.933 \\ \hline
\end{tabular}
\caption{Performance recovery on object detection results by using GNN-based anomaly correction}
\label{tab:postprocessing impact}
\end{table}}

\section{Conclusion}
This paper presents a novel graph-based approach for anomaly detection and correction in annotated visual datasets. By modelling object co-occurrence and spatial relationships using scene graphs, our system effectively identifies mislabeled instances and proposes semantically plausible corrections. Extensive ablation studies reveal the sensitivity of the model to neighbourhood parameters and anomaly severity, highlighting the trade-off between graph density and correction robustness. When applied to the outputs of modern object detectors, our post-processing pipeline leads to substantial improvements in classification accuracy, without retraining the base models. These findings underscore the utility of graph neural networks for quality assurance in large-scale vision datasets and open up avenues for integrating structural reasoning into perception pipelines.

This approach is not limited to egocentric frames and can be broadly applied to enhance object detection performance in any static visual environment. It also serves as a powerful tool for validating manual annotations by identifying labelling inconsistencies and human errors. Beyond correcting mislabelled objects, this framework can be extended to incorporate a bounding box regression module, enabling the recovery of missed detections or refinement of imprecise bounding boxes that fall below confidence thresholds. 

\printbibliography
\end{document}